%% file: atharva.tex
\documentclass[12pt]{article}

\input preamble.tex

\title{Keystroke Dynamics for User Identification}

\author{Atharva Sharma\footnotemark[1]\ \ \ 
Martin  Jure\vvv{c}ek\footnotemark[2]\ \ \ 
Mark Stamp\footnotemark[1]\,\,\footnotemark[3]}

\begin{document}

\symbolfootnotetext[1]{Department of Computer Science, San Jose State University}
\symbolfootnotetext[2]{Faculty of Information Technology, Czech Technical University in Prague}
\symbolfootnotetext[3]{mark.stamp$@$sjsu.edu}

\maketitle

\abstract
In previous research, keystroke dynamics has shown promise for user authentication,
based on both fixed-text and free-text data.
In this research, we consider the more challenging multiclass
user identification problem,
based on free-text data. We experiment with a complex image-like
feature that has previously been used to achieve state-of-the-art authentication results
over free-text data. Using this image-like feature and multiclass 
Convolutional Neural Networks, we are able to obtain a
classification (i.e., identification) accuracy of~0.78 over a set of~148 users. 
However, we find that a Random Forest classifier trained on a slightly
modified version of this same feature yields an accuracy of~0.93.

\section{Introduction}\label{chap:Introduction}

Authentication and intrusion detection are crucial aspects of online security. Conventional 
authentication methods, such as passwords, have limitations, and biometric systems may 
require additional hardware or be unsuitable for specific user groups. Recent research highlights 
the need for accessible and inclusive authentication systems for all users, including
elderly~\cite{BioAuthOlderAdults, BiometricElderly} and disabled individuals~\cite{BiometricDisabled}. 

Keystroke dynamics are a promising means for improved user authentication and identification. 
By analyzing keystroke patterns, a user can be identified based on their distinctive typing style, 
regardless of age or physical ability. Furthermore, keystroke dynamics can aid in detecting an
intruder who has gained unauthorized access to a system, making such it potentially
a useful tool for intrusion detection.

Compared to traditional authentication methods such as passwords, keystroke dynamics offer several benefits. 
First, keystroke dynamics are challenging to break since people tend to have distinctive typing patterns 
that may be difficult to replicate or guess. In contrast, passwords can be compromised through data 
breaches or guessed through trial-and-error. Second, keystroke dynamics can provide a more robust and 
reliable two-factor authentication approach---if an unauthorized user obtains a valid user's login credentials, 
they may still be detected and denied access, due to their failure to mimic the expected
typing characteristics. Also, keystroke dynamics can offer continuous authentication, 
enabling passive, ongoing user-identity verification throughout a session, adding an extra layer 
of intrusion detection. Overall, keystroke dynamics may enable improvements in authentication, 
identification, and intrusion detection.

For the research presented in this paper, we use the so-called Buffalo free-text keystroke 
dataset to study keystroke dynamics.
This dataset was collected by researchers at SUNY Buffalo
and has been widely used in research in this field~\cite{YanSun}.
Here, free-text means that subjects do not type the same thing, which is in
contrast to fixed-text data, where all subjects type the same text. While
fixed text is used to study problems related to improved authentication (typically, via passwords),
free text is useful for studying the intrusion detection problem which, in the context of free text,
is sometimes referred to as continuous authentication.

Both free-text and fixed-text data can be used to study 
the user identification problem. Note that in this context,
for the identification problem we are trying to
determine specifically who is typing, and there may be a very large number
of possible typists. In contrast, for the authentication problem, the typist
claims to be a specific user, and we only need to determine whether
the person typing is the claimed user or not. Consequently, the 
authentication problem can be viewed as a 1-to-1 comparison,
whereas the identification problem is a many-to-one comparison,
and hence the identification problem is inherently more challenging.
In this paper, we consider the user identification problem,
based on the Buffalo free-text dataset.

Free-text and fixed-text datasets have their advantages and drawbacks. 
Free-text datasets, collected while users type naturally without constraints, 
offer a more realistic representation of user behavior, providing a more 
transparent experience for users~\cite{MONTALVAOFILHO20061440}. On the other hand, 
fixed-text datasets, collected under controlled conditions where participants type specific 
words, phrases, or sentences, enable more controlled experiments and easier comparison 
by eliminating variations in text input~\cite{killourhy2009comparing}. Due to the practicality 
and user experience aspects, we have chosen to work with free-text data in this study.
Note that of the various permutations involving free-text or fixed-text for authentication
or identification, the user identification based on free-text data is the most challenging case.
Note also that in this context, identification is synonymous with classification.

Inspired by successful authentication results in prior studies, 
we first consider a feature engineering approach
that originated in~\cite{image_based_keystroke2}, where elementary
features are transformed into a multi-channel image-like transition matrix which is
referred to a Keystroke Dynamics Image (KDI).
Within this matrix, rows and columns denote keyboard keys, while the depth 
signifies distinct feature categories.
We conduct multi-class classification experiments on the~$148$ users in our dataset, 
employing a Convolutional Neural Network (CNN) model trained on the KDI features with 
cutout regularization. To assess the effect of keystroke sequence lengths on our model, 
we experiment with multiple sequence lengths. The CNN model results yield
a respectable accuracy of~0.78. 

We then experiment with classic learning techniques
using a flattened version of the KDIs as our feature vectors.
We find that a Random Forest model trained on these features yields much improved
results, with an accuracy of~0.93 on this inherently challenging user identification problem.
To the best of the authors' knowledge, this is the strongest result to date
for the user identification problem, based on the popular Buffalo free-text dataset.

The remaining paper is organized as follows:
In Section~\ref{chap:background}, we delve into background topics such as the learning 
techniques utilized and the dataset considered in our study. This section also includes a 
review of related prior research. 
Section~\ref{chap:feature_engineering} details the features we employ and, 
specifically, discusses our feature engineering strategy for preparing input data for our 
classification models. In Section~\ref{chap:architecture}, we elaborate on the 
model architectures considered in this paper and discuss hyperparameter tuning. 
Section~\ref{chap:experiments_and_results} encompasses our experiments and provides
an analysis of the results. Lastly, Section~\ref{chap:conclusion} offers a conclusion and 
suggests potential avenues for future research.

\section{Background}\label{chap:background}

Authentication is a fundamental aspect of security systems~\cite{security-engineering}, 
and keystroke dynamics has emerged as a promising method for verifying user identity. 
Unlike traditional authentication methods, keystroke dynamics has the potential to detect 
intruders even after they have gained access to the system, making it a valuable tool for preventing 
security breaches. However, the effectiveness of keystroke dynamics-based systems 
depends on the ability to accurately classify users, based on their typing characteristics.

The more challenging problem of user identification based on keystroke dynamics 
is also of interest, particularly in the context of an intrusion detection system (IDS).
For such a scenario, the use of free text data may be
advantageous, as compared to fixed text~\cite{AhmedAhmed}. Free text data is more 
representative of how users type on a regular basis and is not constrained by a 
pre-determined text input. This may result in more accurate and reliable 
outcomes. Moreover, free text datasets are adaptable to passive monitoring
within an IDS.

Another advantage of keystroke dynamics-based systems is that they can benefit users 
of all ages and those with disabilities, provided that they type to use a 
system~\cite{TouchpadBehavioralBiometrics}. Therefore, this approach can provide 
a more inclusive and accessible method that does not discriminate 
based on age or physical ability.

In summary, keystroke dynamics-based systems may offer a reliable and effective means of 
user authentication and identification, provided that we can accurately and efficiently distinguish 
between users. In this research, we show that even for the inherently challenging identification
problem, it is possible to obtain strong results.

\subsection{Related Work}

Keystroke dynamics is a behavioral biometric that has been extensively studied for user 
authentication and less so for identification. In~1980, Gaines, et al.~\cite{gaines1980authentication} analyzed 
digraph latencies to examine the distinctiveness of typing patterns and found that specific digraphs 
could distinguish right-handed touch typists from one another with~92\%\ accuracy over a 
limited number of users. Following this, in~1990, Bleha, et al.~\cite{bleha1990computer} proposed a 
real-time pattern recognition based approach to classify users. The online verification system 
they developed had a false rejection rate (FRR) of~8.1\%\ in rejecting valid users 
and~2.8\% false acceptance rate (FAR). This work laid the foundation for much of the
subsequent research in this field.

Recently, machine learning techniques have been widely applied in keystroke dynamics. 
Classic machine learning algorithms, such as $k$-Nearest Neighbors ($k$-NN) and 
Support Vector Machines (SVM), 
have yielded promising results in user authentication tasks. However, these methods 
often rely on handcrafted features, which may be less robust and less generalizable to diverse 
user groups and typing scenarios.

An SVM-based method by Giot, et al.~\cite{GiotSVM}, requires only five captures for initial enrollment, 
while Gingrich, et al.~\cite{Gingrichknn} utilize a $k$-NN approach,
resulting in further improvements in efficiency. These approaches offer robust 
and generalizable methods with high accuracy and efficiency compared to previous work that
utilized traditional statistical-based classification algorithms.

Clustering techniques have been employed in the context of keystroke dynamics to group similar 
users or typing patterns, and to identify potential outliers. Revett, et al.~\cite{kmeans_keystroke}
have demonstrated that $K$-Means clustering can yield
useful results, achieving an authentication accuracy of~96.20\%. In addition, 
clustering techniques can also be used as data analysis tool. For example, 
Robinson, et al.~\cite{hierarchical_keystroke} use
hierarchical clustering to evaluate the effect of hold times on the 
homogeneity of valid user timing vectors. This use of hierarchical clustering helped to 
establish the relative homogeneity of valid user timing vectors and improve the accuracy 
of subsequent experiment.

Clustering can also be applied to keystroke dynamics for the purpose of detecting account sharing. 
Hwang, et al.~\cite{keystrokes_account_sharing}, show that user's keystroke patterns 
forms a distinctive clusters in Euclidean space and the number of shared accounts can then
be estimated by the number of clusters. The optimal number of clusters is estimated using 
a Bayesian model-selection framework, and the results show a~2\% false alarm rate, a~2\% miss rate, 
and a~93\% accuracy. 

Clustering methods such as Expectation Conditional Maximization (ECM) have also been combined 
with other approaches, including Extreme Learning Machines (ELM), to improve accuracy and stability. 
ELM is a single hidden layer feedforward network that is extremely fast to train and 
achieves good generalization performance for some problems.
Sriram, et al.~\cite{ecm-elm} used a clustering-based, semi-supervised ECM-ELM
approach to achieve an accuracy of~87\% with the popular CMU Keystroke Dataset.

Deep learning techniques for analyzing keystroke dynamics have shown promise in recent studies, 
with CNN being employed to achieve notable results. A novel approach 
by Liu and Guan~\cite{cnn_keystroke1} involves converting keystroke data into images-like features, 
which allows for the mining of spatial information and results in an accuracy of~96.8\%,
with an FAR of~0.04\%. In contrast, Piugie, et al.~\cite{cnn_keystroke2} concentrate on using 
deep learning for passphrase-based user authentication, and surpass the performance of 
state-of-the-art methods in terms of the Equal Error Rate (EER).

As researchers explore various deep learning model architectures for keystroke dynamics authentication, 
recent studies have investigated the application of recurrent neural networks. For example,
an architecture based on a hybrid CNN and Gated Recurrent Unit (GRU) is proposed and analyzed 
by Lu, et al.~\cite{rnn_keystroke1}, while Mhenni, et al.~\cite{LSTM_KEYSTROKE1} examine 
the use of Long Short-Term Memory (LSTM) and Bidirectional Long Short-Term Memory (BiLSTM) 
architectures. Both papers illustrate the potential of deep learning models in this domain. In particular,
Mhenni, et al., show that BiLSTM outperforms LSTM, 
achieving an accuracy of~86\% and~71\% for the GREYC-2009 and WEBGREYC databases, 
respectively; in comparison their LSTM model has an accuracy of~68\% and~53\% over these same
datasets.

The research presented in this paper is motivated by previous work
involving image-like data structures for keystroke 
data~\cite{image_based_keystroke2, image_based_keystroke1}. 
These image-like representations can leverage the powerful capabilities of CNNs, which are known 
for their success in image classification tasks. In this context, the work of 
Li, et al.~\cite{image_based_keystroke2} is particularly relevant, as it introduced a unique 
Keystroke Dynamic Image (KDI) that lead to improved state-of-the-art
results, as compared to previous work. We consider this same KDI image-like 
feature in our multiclass experiments.

In summary, the related work in the field of keystroke dynamics spans a wide range of techniques 
and methodologies, including classic machine learning, deep learning, feature engineering, 
threshold-based techniques, clustering, and various ensembles. Building upon this rich body of research, 
the current study aims to advance the state-of-the-art, particularly within the relatively
neglected area of user identification.

\subsection{Dataset}

For our experiments, we are use a free-text keystroke dataset collected by researchers
at SUNY Buffalo~\cite{YanSun}, which is referred to as the Buffalo Keystroke Dataset in the literature,
or more simply, the Buffalo dataset.
This dataset is a collection of free-text keystroke dynamics data obtained from~148 
research participants. The participants were asked to complete two typing tasks in a 
laboratory setting over the course of three separate sessions. The first task involved 
transcribing Steve Jobs' Commencement Speech, split into three parts, while the second 
task included free-text responses to questions. To ensure the generalizability, 
there was a~28-day interval between each session.

Out of the~148 participants, 75 completed the typing test with the same keyboard 
across all three sessions, while the remaining~73 participants used three different 
keyboards in each session. The dataset contains the timestamp of key presses (key-down) 
and key releases (key-up), organized in a tabular format with three columns; the first column 
indicates the key, the second column denotes whether the event is a key-press or key-release, 
and the third column records the timestamp of the event. The dataset includes information about 
the gender of each participant, and on average, each participant has a total of 
over~$17{,}000$ keystrokes across their three sessions.
The Buffalo Keystroke Dataset has been widely used in the research literature.

\subsection{Machine Learning and Deep Learning Algorithms}

Despite the rapid growth of neural networks, classic machine learning algorithms remain 
competitive in the field of keystroke dynamics. Such algorithms are based on statistical 
and mathematical techniques, and have been used with success for many years in 
various fields. Among deep learning techniques, we consider 
Convolutional Neural Networks.

\subsubsection{Support Vector Machines}

Support Vector Machine (SVM)~\cite{SVM} is a powerful supervised machine learning technique,
which has its theoretical foundation solidly rooted in computational and mathematical principles, 
SVM is designed to identify a hyperplane in an $N$-dimensional space that can accurately
separate labeled data points into their respective classes. The algorithm aims to maximize the 
minimum distance, or ``margin,'' between the hyperplane and the data. SVM is generally recognized 
for its practical effectiveness, as it can efficiently handle large and complex datasets. It has been 
used in a wide range of fields, including image classification, text classification, and bioinformatics.

\subsubsection{Random Forest}

Random Forest classifiers consist of ensembles of decision trees.
During training, a Random Forest uses a 
divide and conquer strategy by sampling small subsets of the data and features, 
with a simple decision tree constructed for each such subset. The Random Forest 
classification is based on the predictions of its component 
decision trees, usually using a simple voting strategy~\cite{biau2016random}. 
Important hyperparameters in a Random Forest include the number of estimators
(i.e., decision trees), maximum features (maximum number of features to sample in
any one decision tree), among others.

%

\subsubsection{Convolutional Neural Network}

CNNs~\cite{CNN} are a specialized type of neural network that utilize convolution 
kernels to deal with local information, often from image-like data. 
Unlike traditional neural networks, CNNs share weights at different locations, 
resulting in more efficient and shift-invariant models with fewer parameters. 
Their multi-layer convolutional architecture enables them to extract information at different 
resolutions in computer vision tasks, making them ideal for image processing. CNNs can analyze 
images and extract important features, such as edges, shapes, and textures, in a highly effective manner. 
Additionally, the use of convolution kernels in CNNs enables the network to learn spatial features, 
such as orientation and scale, which is especially useful in image recognition tasks. CNNs have proven 
to be highly effective in a surprisingly wide variety of applications, including object recognition, 
face recognition, and image classification. CNNs have also been successfully applied to non-image data, 
such as audio and text.

Dropout regularization~\cite{srivastava2014dropout} is widely used to prevent overfitting 
in feedforward neural networks. However, this approach is less effective in convolutional layers due to their 
shared information and lower parameter count. To overcome this limitation, Cutout regularization 
is used~\cite{Cutouts}. As the name suggests, Cutout regularization consists of blocking out parts of the input image
at various stages in the training process. This forces the model to focus on areas of the image
that might otherwise be ignored during training, resulting in a more robust model. Cutouts also improve
a model's ability to generalize and perform well with limited training data. Overall, Cutout is a 
versatile and effective technique for image analysis that can enhance the performance of CNNs.

\section{Feature Engineering}\label{chap:feature_engineering}

As mentioned above, we use the Buffalo Keystroke Dataset, a free-text dataset with limited information. 
Feature engineering is critical to our analysis, as we will be exploring image-like features that are 
derived from the features existing in the dataset. These features capture the timing information 
of individual keystrokes and their relationships to other keystrokes, allowing us to build a detailed sequence 
of keystrokes for each user. By carefully engineering these features, we hope to gain 
additional insight into how keystroke dynamics can be successfully used as a biometric for user 
identification and authentication.

\subsection{Keystroke Features}\label{section:keystroke_features}

Keystroke dynamics datasets often provide two kinds of features, namely, 
time-based information and pressure-based information. 
Both types of features can provide valuable insights into typing behavior, 
but pressure-based features are not available on many modern keyboards. 
Therefore, our research will focus solely on time-based information. 

\begin{figure}[!htb]
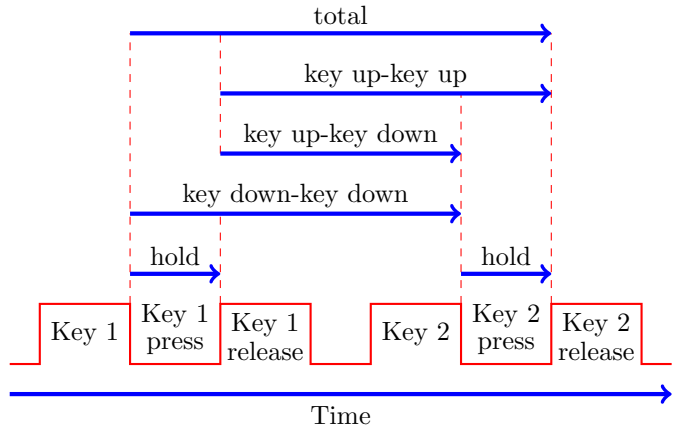

\centering
\input figures/kdFeatures.tex
\caption{Time based features}\label{fig:time_based_features}
\end{figure}

The data in the Buffalo Keystroke Dataset can be understood by examining the
following five time-based features, which are depicted in Figure~\ref{fig:time_based_features}. 
\begin{itemize}
\item Duration: The time that the user holds a key in the down position
\item Down-down time (DD-time): The time between the press of a key and the press of the subsequent key
\item Up-down time (UD-time): The time between the release of a key and the press of the subsequent key
\item Up-up time (UU-time): The time between the release of a key and the release of the subsequent key
\item Down-up time (DU-time): The time between the press of a key and the release of the subsequent key
\end{itemize}

Note that for any two consecutive keystroke events, denoted Key~1 and Key~2, 
six features can be extracted: duration of Key~1, duration of Key~2, DD-time, UD-time, 
UU-time, and DU-time. By carefully analyzing these features, we hope to gain insights 
into the unique patterns of typing behavior exhibited by individual users and 
determine how these patterns can be used for user identification.

\subsection{Keystroke Sequence}

A keystroke sequence refers to the entire series of keystrokes entered by a user. To better analyze 
these sequences, they are often divided into smaller subsequences. Each subsequence can be 
viewed as an independent keystroke sequence from the same user.

In our research, we will experiment with different lengths of keystroke subsequences, which
we treat as a hyperparameter of the system. A longer keystroke sequence can provide more information, 
but it can be more resource-intensive to process, and also delays the analysis until the
sequence has been collected. Shorter subsequences may not capture enough information and 
thereby result in decreased accuracy. Therefore, we will carefully select the length of keystroke 
subsequences to ensure that they are optimized for accuracy while being mindful of resource 
usage. Additionally, the length of the subsequences will impact the creation of image-like 
features that we will be using in our analysis. By selecting the optimal length of keystroke 
subsequences, we aim to improve the accuracy and practicality of our results.

\subsection{Keystroke Data Image}

In the previous section, we discussed dividing the keystroke sequence into multiple subsequences. 
As discussed in Section~\ref{section:keystroke_features}, there are six types of timing features. 
Therefore, for a subsequence of~$N$ keystrokes, we can determine $6(N - 1)$ features from consecutive 
pairs of keystrokes. Repeated pairs are averaged and treated as a single pair. For instance, 
a subsequence of length~50 would yield at most $6\cdot 49 = 294$ features. 
We consider each keystroke subsequence as an independent input sequence for 
the corresponding user. In this section, we discuss a feature engineering structure 
originally developed in~\cite{image_based_keystroke2},
that enables us to effectively organize these features.

\begin{figure}[!htb]
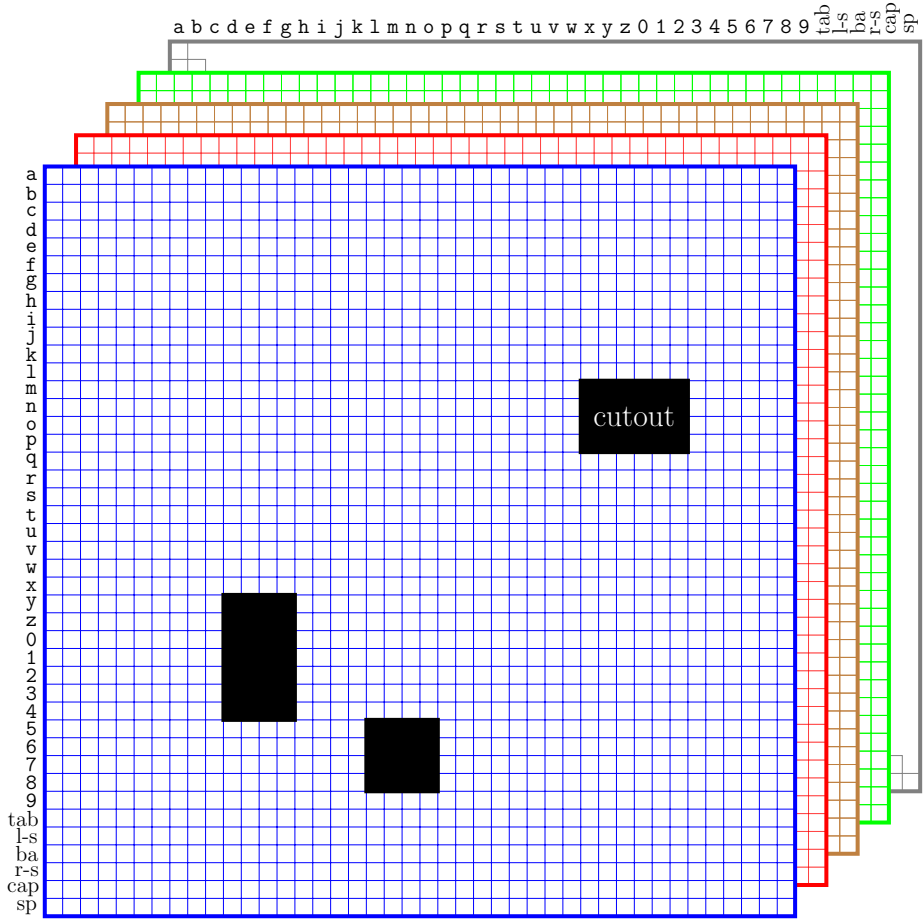

\centering
\adjustbox{scale=0.95}{
\input figures/keystroke_img.tex
}
\caption{Image-like features from keystrokes}\label{fig:keystroke_dynamics_image}
\end{figure}

The features UD-time, DD-time, DU-time, and UU-time are determined by consecutive keystroke events. 
We organize these four features into a transition matrix with four channels, inspired by the structure 
of~RGB images, which have a depth of three (R, G, and B channels). Each row and column in 
our four-channel~$n\times n$ feature matrix corresponds to a key on the keyboard, with each channel 
representing one kind of feature. We organize these into transition matrices as shown in Figure~\ref{fig:keystroke_dynamics_image}, which we refer to as 
the KDI feature.

For example, in the first channel of the matrix, the value at row~$i$ and column~$j$ refers to 
the UD-time between any key presses of~$i$ followed by~$j$ within the current observation 
window. We add the final feature, duration, as a diagonal matrix to the transition matrix, 
creating a fifth channel. If a key or key-pair is pressed more than once, we use the average 
duration for that key or key-pair. In this fifth channel, only diagonal locations have values because 
the duration feature is relevant for one key at a time. We can use this transition matrix as an image 
input for machine learning models. To avoid sparsity in the transition matrix, we only consider 
time-based features for the following~42 most common keystrokes. 

\begin{enumerate}
\item The~26 English characters (A-Z)
\item The~10 Arabic numerals (0--9)
\item The following~6 meta keys: space, back, left-shift, right-shift, tab, and capital
\end{enumerate}
Thus, the shape of the transition matrix is~$42 \times 42 \times 5$, with the five channels as described above. 

In order to prevent overfitting in our CNN, we make use of cutout regularization,
as discussed above. 
The dark blocks in Figure~\ref{fig:keystroke_dynamics_image} represent cutouts.

\section{Model Architectures}\label{chap:architecture}

In this section, we outline the learning architectures used in
our experiments. We also discuss hyperparameter tuning for each of
our models.

\subsection{Multiclass CNN}

\begin{figure}[!htb]
\centering
\includegraphics[width=0.75\textwidth]{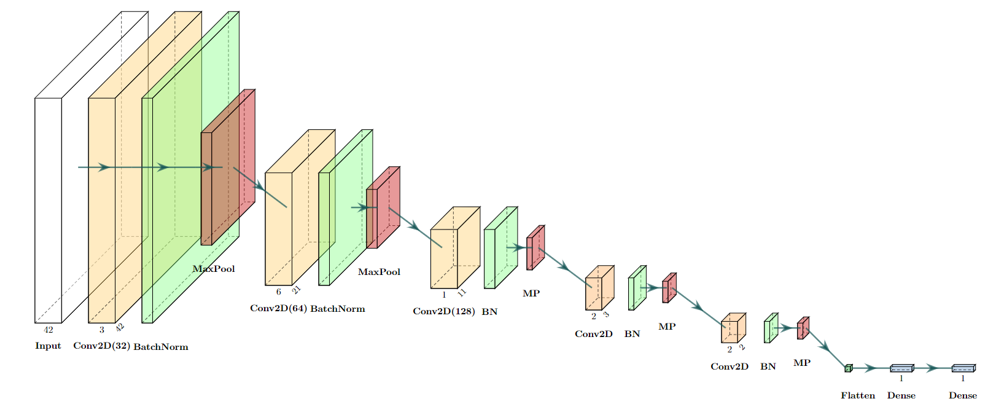}
\caption{Architecture of sequential CNN}
\label{fig:multiclass_cnn}
\end{figure}

The architecture shown in Figure~\ref{fig:multiclass_cnn} is a sequential CNN-based neural network. 
The input shape of the model is~$(5, 42, 42)$, indicating that the input is a~\hbox{3-D} array with a depth of~5 
and a width and height of~42. The model includes five convolutional layers, each followed by a 
batch normalization layer and a max pooling layer. The first convolutional layer has~32 filters of size~$(5,5)$, 
while the subsequent convolutional layers have~64, 128, 256, and~256 filters of size~$(3,3)$, 
respectively. All convolutional layers use the Rectified Linear Unit (ReLU) activation function. 
The max pooling layers have a pool size of~$(2,2)$ and a stride of 2, ensuring that the output 
size of each layer remains the same.

After the five convolutional layers, the model includes a flatten layer, followed by two fully 
connected layers. The first fully connected layer has~128 units with the ReLU activation 
function. The final output layer has~148 units with the softmax activation function. 
The model is compiled using categorical crossentropy as the loss function, 
the adam optimizer, and we use accuracy as the evaluation metric.

To identify the best combination of hyperparameters, we employed a grid search 
over reasonable values of various parameters. The hyperparameters tested for our CNN are 
given Table~\ref{hyperparameters}, where the selected values are in boldface We use
these hyperparameters in all CNN models discussed 
in Section~\ref{chap:experiments_and_results}, below.

\begin{table}[!htb]
\caption{Hyperparameter tuning for multiclass CNN}\label{hyperparameters}
\centering
\adjustbox{scale=0.85}{
\begin{tabular}{c|c}\midrule\midrule
Parameter &  Values \\ \midrule
Number of epochs & 10, \textbf{20}, 30, 40  \\
Learning rate & 0.1, \textbf{0.01}, 0.001, 0.0001  \\
Optimizer & \textbf{Adam}, SGD, SGD with Momentum  \\
Learning schedule & StepLR, \textbf{reduceLROnPlateau}\\ \midrule\midrule
\end{tabular}
}
\end{table}

Note that our best model uses reduceLROnPlateau (from Keras) callback to dynamically 
reduce the learning rate when the model was unable to improve during training. Also,
we utilize earlyStopping (also from Keras) callback to halt training if the model 
showed signs of overfitting. Our experimental results indicated that the model tended to 
overfit after the~$20\thth$ epoch, which confirmed the results of our grid search.

\subsection{SVM Classifier}

We consider the classic SVM learning technique. 
As discussed above, for our SVM classifier we use the flattened
KDI, based on a one-hot encoding to key events. 
These features are then standardized to have zero mean and unit variance.

The SVM classifier used for multiclass classification is a one-vs-one (OVO) classifier, 
which trains a separate SVM for each pair of classes. 
Since this is costly to train, we restrict our attention to the~48 users that
are most difficult to classify using our CNN.
This requires that we train a total of~${48 \choose 2} = 1128$ SVM classifiers. 
To train the SVM classifier, the dataset is split into training and testing sets using 
stratified random sampling, with an~80-20 split, that is, 80\% of the data is used for 
training and~20\% for testing. To further reduce the training time 
we use the following hyperparameters: $C = 1$, 
$\mbox{kernel} = \mbox{rbf}$, and $\gamma=\mbox{scale}$.

\subsection{Random Forest Classifier}

We also train and test Random Forest classifiers. 
As with our SVM model, for our Random Forest classifier we use the
flattened KDI, based on a one-hot encoding to key events. 
The Random Forest hyperparameters
tested for this model are listed in Table~\ref{hyperparameters-RF}, with the
values selected in boldface.

\begin{table}[!htb]
\caption{Hyperparameter tuning for Random Forest}\label{hyperparameters-RF}
\centering
\adjustbox{scale=0.85}{
\begin{tabular}{c|c}\midrule\midrule
Parameter &  Values \\ \midrule
n\_estimators & 100, 500,~\textbf{1000}  \\
max\_features &~\textbf{auto}, sqrt  \\
min\_samples\_split & 2,~\textbf{5}  \\
min\_samples\_leaf & 1,~\textbf{2}\\ \midrule\midrule
\end{tabular}
}
\end{table}

As with our SVM classifier, we initially restrict our Random Forest to the~48 most
challenging to identify users. However, given the strong results that we obtain,
and since there is no significant efficiency issue when training on a larger dataset,
we also train and test this Random Forest on the entire set of~148 users.

\section{Experiments and Results}\label{chap:experiments_and_results}

In this section, we first discuss our experimental design.
Then we present our experimental results, and provide some discussion of these results.

\subsection{Experiment Strategy}

We train three multiclass CNN classifiers over all~148 users,
based on the KDI data structure and keystroke subsequence lengths 
of~50, 75 and~100. Once we establish our best model, 
we generate the confusion matrix, and sort based on the diagonal
(i.e., true positive) elements. This enables us to split the users into three categories, 
namely, those that are easiest to classify, those that are of moderate difficulty to classify,
and those that are the most difficult to classify.

We then apply classic machine learning techniques to 
difficult-to-classify users, based on a flattened KDI, that is,
we convert the~$5\times 42\times 42$ KDI into
a feature vector of length~$5\cdot 42\cdot 42 = 8820$.
The performance of the classic techniques on these challenging cases
leads us to further analyze the best of the models over the entire dataset.

\subsection{Metrics}

We use accuracy as the primary means of measuring the quality of our results.
We also present confusion matrices to better visualize the distribution of correct and incorrect 
predictions across all classes, and to distinguish users, based on the difficulty of 
correct classification. 


The accuracy of a binary classifier is simply the number of correct classifications
divided by the total number of classifications, that is, 
$$
\mbox{Accuracy} = \frac{\mbox{TP} + \mbox{TN}}{\mbox{TP} + \mbox{TN} + \mbox{FP} + \mbox{FN}},
$$
where TP is the number of true positive samples (samples correctly classified as positive), 
TN is the number of true negative samples (samples correctly classified as negative), 
FP is the number of false positive samples (samples incorrectly classified as positive), 
and FN is the number of false negative samples (samples incorrectly classified as negative).
In a multiclass classification problem, accuracy is the proportion of correctly classified 
samples to the total number of samples in the dataset. We can calculate the accuracy for 
multiclass classifier as 
$$
\mbox{Accuracy} = \frac{\displaystyle\sum_{i=1}^{n} \mbox{TP}_i}{M},
$$
where~$\mbox{TP}_i$ represents the number of samples of class~$i$ that are correctly 
classified and~$M$ is the total number of samples in the dataset.

In a confusion matrix, each row and column corresponds to a class in the dataset.
We follow the convention that the rows represent the actual classes of the samples, 
and the columns represent the predicted classes. To determine the 
accuracy for class~$i$, we simply divide the~$i^{\thth}$
diagonal element by the sum of the elements in row~$i$. As noted above, 
the overall accuracy is the sum of all diagonal elements, divided by the sum of all
elements in the matrix.

\subsection{Multiclass CNN Experiments}\label{section:results_multiclass}

As discussed above,
we determined our CNN hyperparameters via a grid search over the 
number of epochs, learning rate, optimizer, and learning schedule callbacks. 
We found that training for~20 epochs, with a learning rate of~0.01, along with adam 
and reduceLROnPlateau as optimizer and callback, respectively, yielded the best results. 
We also experimented with different architecture of the model itself,
and settled on a model with five convolutional layers, where each convolutional layer 
is followed by batch normalization and max pooling layers, as illustrated in
Figure~\ref{fig:multiclass_cnn}, above. 

\begin{figure}[!htb]
\centering
\begin{tabular}{cc}
\includegraphics[width=0.425\textwidth]{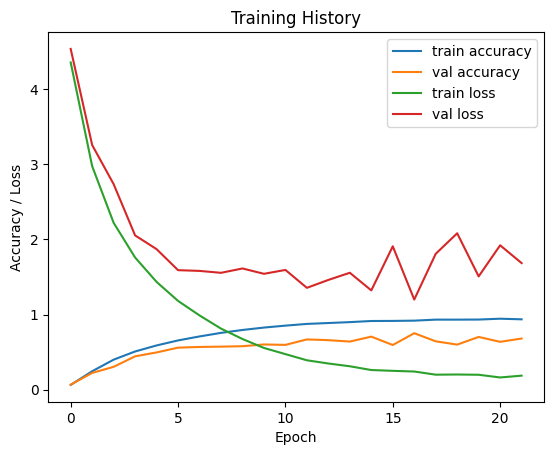}
&
\includegraphics[width=0.425\textwidth]{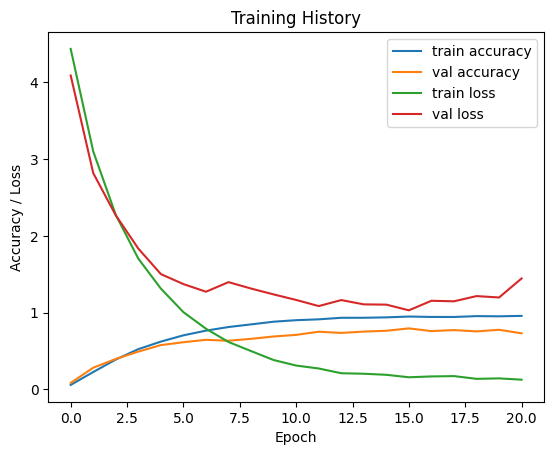}
\\
\adjustbox{scale=0.9}{(a) Keystroke length~50}
& 
\adjustbox{scale=0.9}{(b) Keystroke length~75}
\\ \\[-1ex]
\multicolumn{2}{c}{\includegraphics[width=0.425\textwidth]{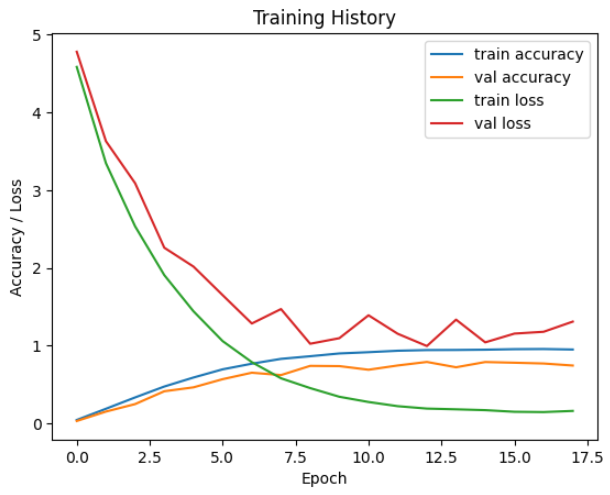}}
\\
\multicolumn{2}{c}{\adjustbox{scale=0.9}{(c) Keystroke length~100}}
\end{tabular}
\caption{Training of models}\label{fig:50_75_100_graphs}
\end{figure}

After determining the hyperparameters and model architecture, we 
experimented with the keystroke sequence length for the KDIs. 
The model with keystroke length~50 shows signs of overfitting,
as can be seen from the Figure~\ref{fig:50_75_100_graphs}(a). 
On the other hand, models which were trained on keystrokes with length~75 and~100 
are more robust against overfitting, as can be seen in 
Figures~\ref{fig:50_75_100_graphs}(b) and~(c), respectively---for both
of these cases, the validation loss is continuously dropping and both validation 
and training accuracies are steadily climbing. 

The comparative analysis of training, testing and validation accuracies for keystrokes 
of length~50, 75 and~100 respectively is shown in Table~\ref{Accuracies}. 
Since the model trained on keystroke sequence~100 gave us the best test and validation
accuracies, we will extract the confusion matrix from this model
for additional experiments in the next section.

\begin{table}[!htb]
\caption{Accuracy as a function of keystroke sequence length}\label{Accuracies}
\centering
\adjustbox{scale=0.85}{
\begin{tabular}{c|cccc}\midrule\midrule
\multirow{2}{*}{Length} & \!\!\!\!\! & \multicolumn{3}{c}{Accuracy} \\ \cline{3-5} \\[-2.25ex]
    & \!\!\!\!\! & Train & Test & Validation\\ \midrule
50 & \!\!\!\!\! & 0.9074  & 0.6700 & 0.5800\\
75 & \!\!\!\!\! & 0.9500 & 0.7400 & 0.7300 \\
100 & \!\!\!\!\! & 0.9700  & 0.7900 & 0.7800 \\ \midrule\midrule
\end{tabular}
}
\end{table}

\subsubsection{CNN Confusion Matrix}

\begin{figure}[!htb]
\centering
\includegraphics[width=0.7\textwidth,height=0.45\textwidth]{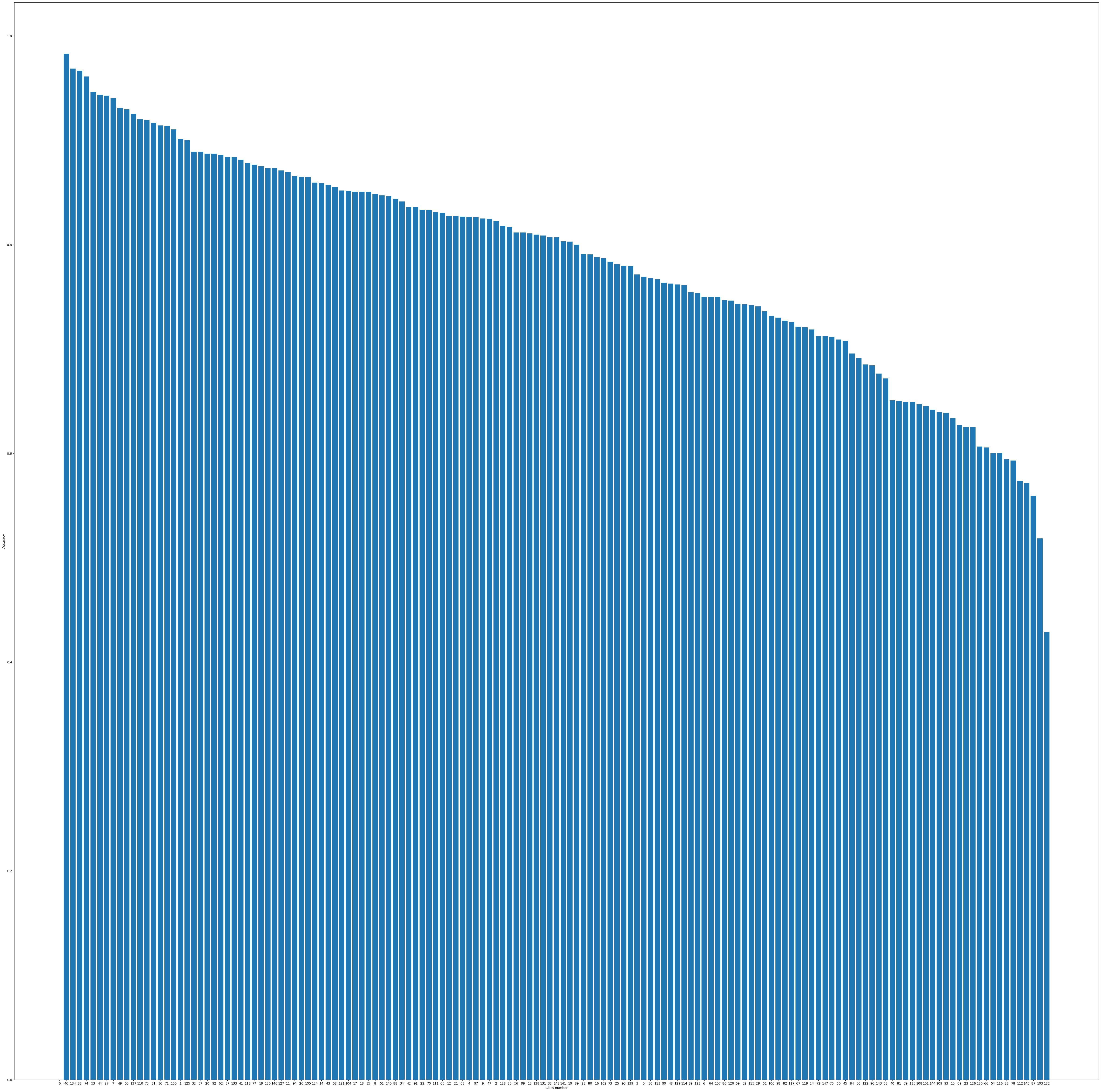}
\caption{Bar graph for~148 users in dataset}
\label{fig:100_histogram}
\end{figure}

As established in Section~\ref{section:results_multiclass}, 
the model trained on keystroke length~100, provides the best results. 
For this keystroke length~100 model, a bar graph of the accuracy for each
user is is given in Figure~\ref{fig:100_histogram}, where we have sorted by
accuracy. 

\begin{figure}[!htb]
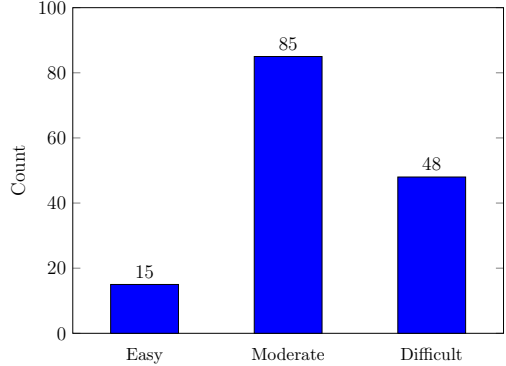

\centering
\adjustbox{scale=0.9}{
\input figures/barClusters.tex
}
\caption{Number of users in each cluster}
\label{fig:3_clusters}
\end{figure}

Next, we use the bar graph in Figure~\ref{fig:100_histogram} to partition the users
into three subsets, based on the accuracy attained when identifying them 
using a CNN trained on the KDI features. 
We use the ``slope'' of the bar graph to identify these subsets. 
A slight ``elbow'' in the slope occurs after about a fifth of the users, 
and another is towards the last fourth of the users. Based on these observations, 
we establish two accuracy thresholds for authenticating users.  
Those users who are classified with~0.90 or greater accuracy,
we consider relatively easy-to-authenticate, while those who are classified
at accuracies below~0.75 are deemed the difficult-to-classify subset, while all of those
in between these two thresholds are referred to as moderate-to-classify. 
The number of users in each of these subsets
can be found in Figure~\ref{fig:3_clusters}.
We further analyze the difficult-to-classify subset in the next section.

\subsection{Experiments on Difficult-to-Identify Users}

In the previous section, we categorized~48 of the users as difficult to identify
using a CNN trained on KDI features. Here, we consider additional experiments on this 
subset of users, to see if we can improve on the poor results for these users.
Specifically, we apply Support Vector Machines, 
Decision Trees, and Random Forest, based on the flattened KDI features. 
For the~48 users that comprise the difficult-to-identify subset, 
we achieve the accuracies listed in 
Table~\ref{Accuracies-HTI-flattened-images}.

\begin{table}[!htb]
\caption{Accuracy of models for difficult-to-identify users}
\label{Accuracies-HTI-flattened-images}
\centering
\adjustbox{scale=0.85}{
\begin{tabular}{c|c}\midrule\midrule
Model & Test accuracy\\ \midrule
SVM & 0.50 \\
Decision Tree & 0.88 \\
Random Forest & 0.92 \\ \midrule\midrule
\end{tabular}
}
\end{table}

\begin{figure}[!htb]
\centering
\includegraphics[width=0.825\textwidth]{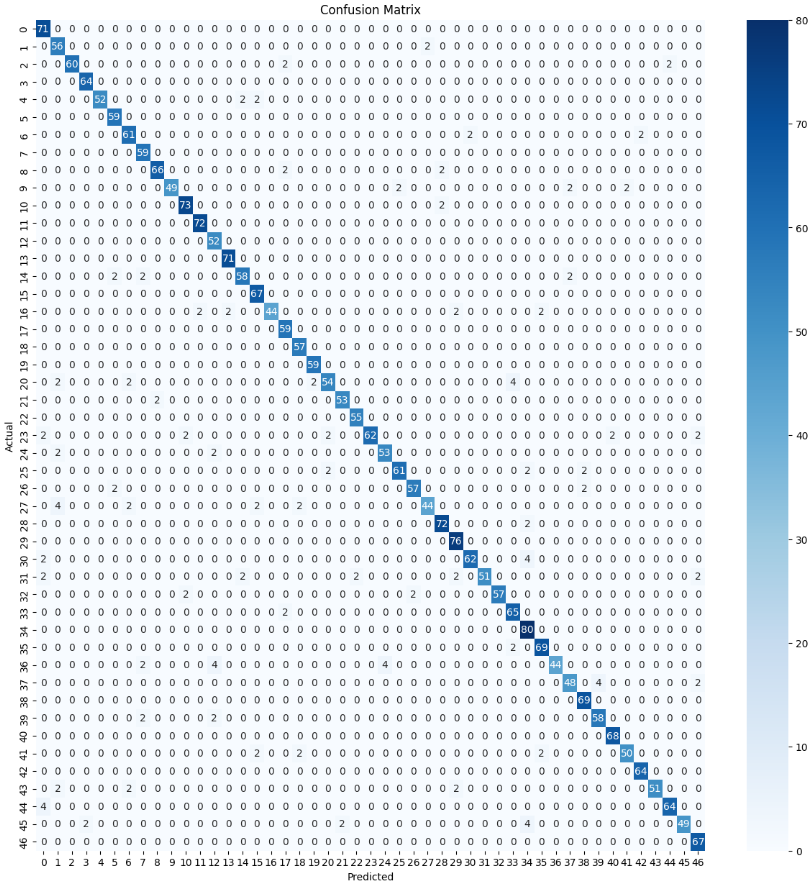}
\caption{Random Forest confusion matrix for difficult-to-identify users}
\label{fig:cnfn_mat_rf}
\end{figure}

\begin{figure}[!htb]
\centering
\includegraphics[width=0.8\textwidth]{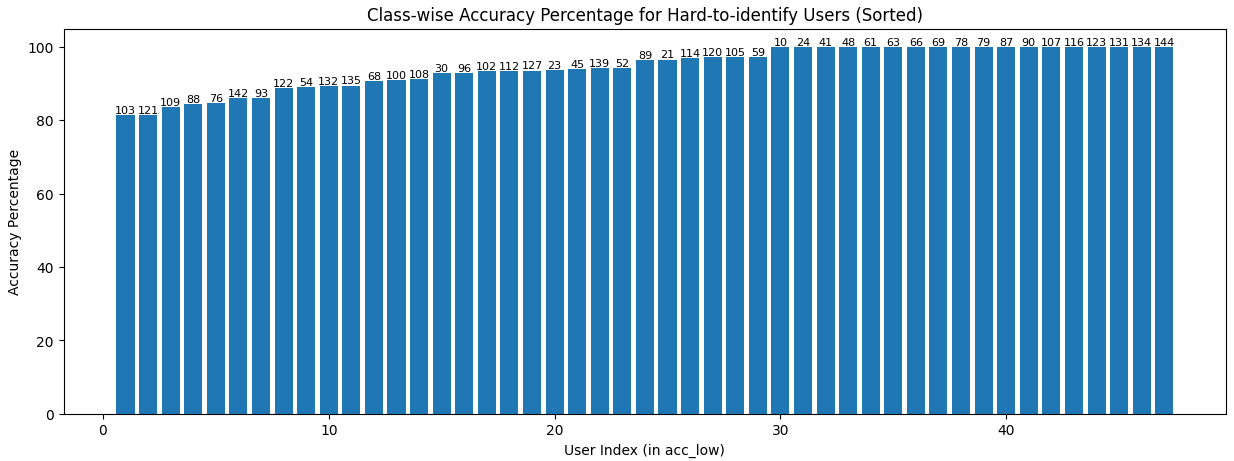}
\caption{Random Forest results for difficult-to-identify users}
\label{fig:hist_rf}
\end{figure}

The confusion matrix for best of these experiments, namely, the Random Forest model,
appears in Figure~\ref{fig:cnfn_mat_rf}. 
In Figure~\ref{fig:hist_rf}, we provide a bar graph of per-user accuracies, sorted in descending order.
These graphs serve to reinforce the strong results that we have obtained classifying the
most challenging users.

\subsection{Random Forest Model for all Users}

Our surprisingly strong results on the difficult-to-classify users
lead us to test the Random Forest model trained on the flattened KDI features 
over the entire dataset of~148 users. We find that the accuracy 
in this case is~0.93.

\begin{figure}[!htb]
\centering
\includegraphics[width=0.65\textwidth]{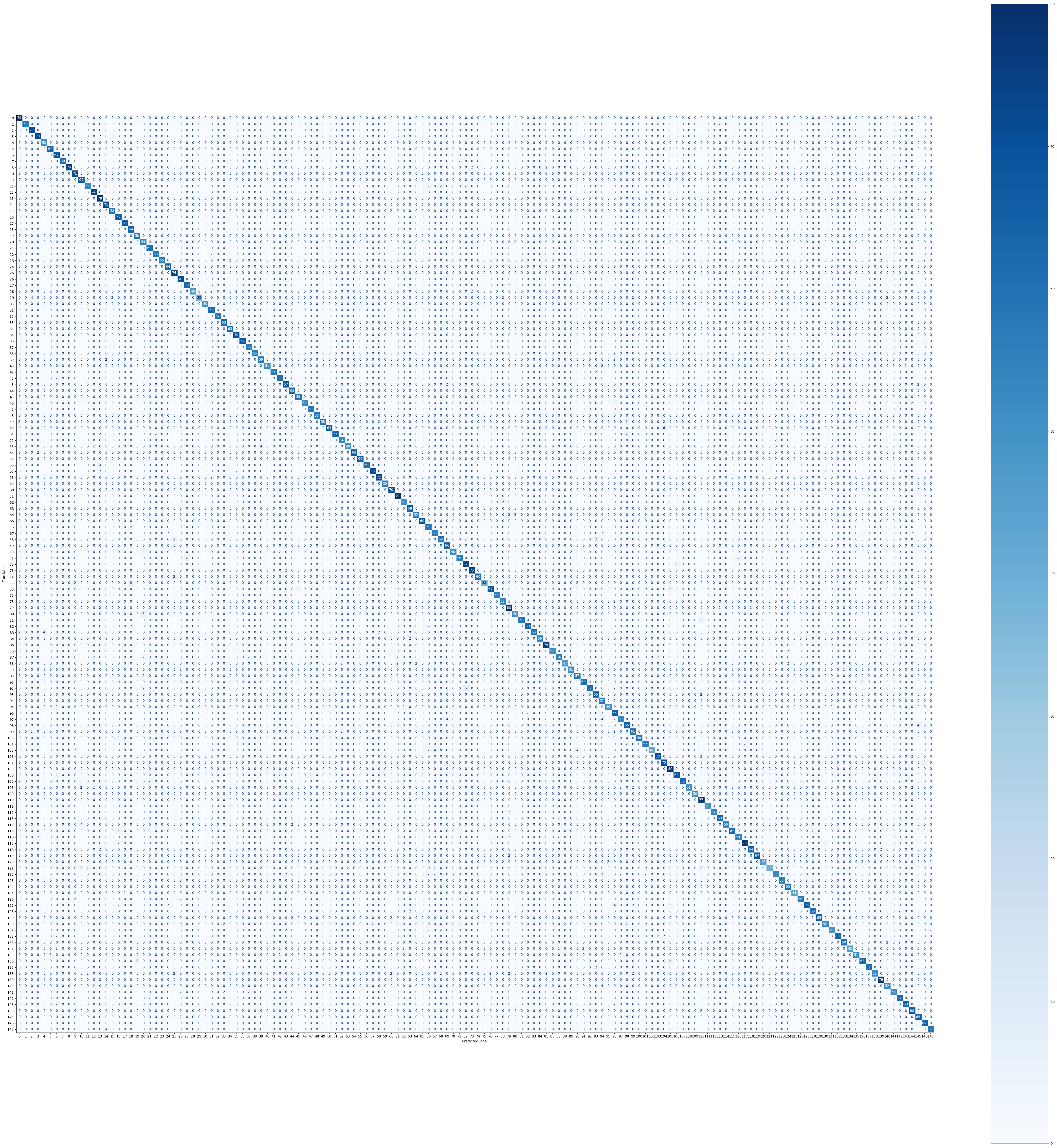}
\caption{Random Forest confusion matrix for all 148 users}
\label{fig:cnfn_mat_rf_complete}
\end{figure}

\begin{figure}[!htb]
\centering
\includegraphics[width=0.8\textwidth,height=0.2\textwidth]{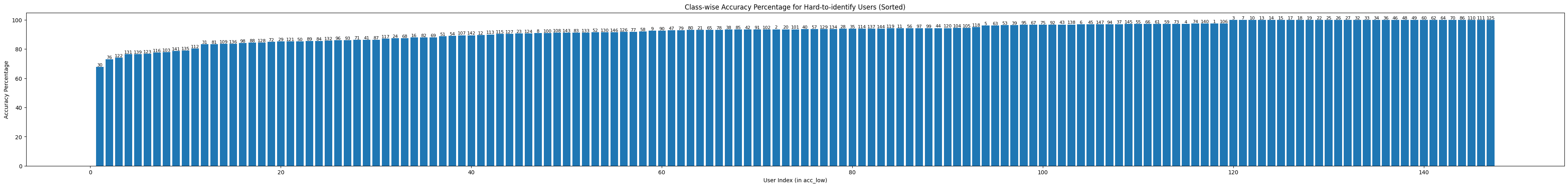}
\caption{Random Forest results for all~148 users}
\label{fig:hist_rf_complete}
\end{figure}

Figure~\ref{fig:cnfn_mat_rf_complete} shows the confusion matrix 
for our Random Forest model for all~148 users. This outcome signifies a 
substantial improvement over the original multiclass CNN that was 
trained on the 5-channel KDI, as the CNN model only achieved an 
accuracy of~0.78.
Figure~\ref{fig:hist_rf_complete} displays the sorted per-user 
accuracy bar graph for our Random Forest model. 

\begin{figure}[!htb]
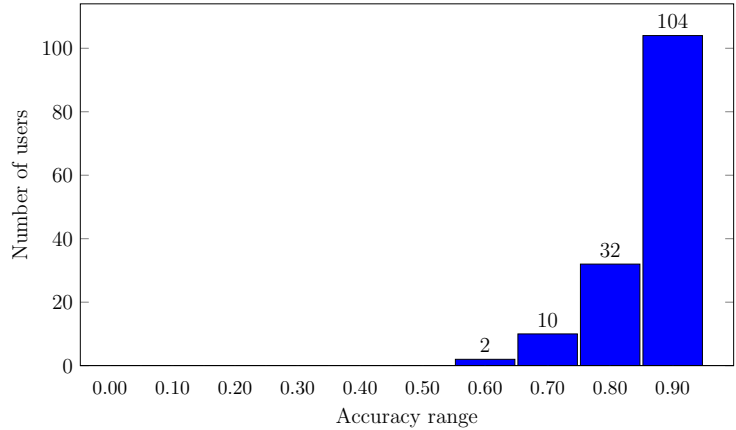

\centering
\input figures/barRange.tex
\caption{Random Forest accuracy ranges for all~148 users}
\label{fig:class_wise_rf_complete}
\end{figure}

Figure~\ref{fig:class_wise_rf_complete} shows the number of users for each classification decile for
our Random Forest model. We find that only a small fraction of users, specifically~8, 
now have an accuracy of~0.75 or lower. In addition, 27 users have classification 
accuracies in the range of~0.75 to~0.90, while the vast majority of users, 
109 to be precise, are classified with an accuracy of at least~0.90.

These results underscore the success of the Random Forest model 
trained on the flattened KDI features. As far as the authors are aware, this
Random Forest model yields the best identification results yet achieved for 
the~148 users in the Buffalo dataset.

 \section{Conclusion}\label{chap:conclusion}

In this research, we expanded on previous work on user identification 
based on keystroke dynamics. We used an image-like data structure and obtained good results
for the identification problem, based on a Convolutional Neural Network. While using this keystroke image-like
features with binary CNN classifiers is not new, one innovative aspect of the research in this paper
lies in the application of a multiclass CNN for identifying users, which enabled us to 
categorize users into easy-, moderate-, and difficult-to-identify subsets. This approach 
enabled us to focus extra attention on the users that are most difficult to identify. 

When experimenting with the most challenging cases, we discovered that a Random Forest
trained on a flattened KDI feature yielded surprisingly strong results.
Even more surprising, testing this same model over the entire dataset yielded 
much better results that our multiclass CNN model.

Future research could explore incorporating additional features, such as digraph and trigraph 
latencies, or even other biometric data, to improve model performance.
Also, the Buffalo keystroke dataset that we employed for our experiments 
was created using mechanical keyboards. It would be valuable to 
obtain keystroke data from mobile devices and apply a similar analysis to that data.
The dynamics of touch-based interactions likely differ substantially from those of traditional 
mechanical keyboard input.  

Another potential area of future research is the development of an efficient strategy for 
adding new users to an existing keystroke dynamics-based
authentication or identification system. Currently, incorporating new users 
into a multiclass model generally requires retraining the entire model, which may not be practical in 
real-world scenarios. To overcome this issue, we could explore methods for 
determining the cluster that a user's keystroke patterns most closely match. It might then be 
possible to achieve good results in a two-stage process, where users are first
assigned to a cluster, and subsequently distinguished from the other users in their cluster.
By assigning a user to the nearest cluster, we could avoid the problem of retraining a
multiclass model for all users. 
This approach could facilitate the seamless integration of new users into the system,
while maintaining the efficiency and accuracy of the identification or authentication process.

In summary, we have achieved strong results for the challenging problem of 
user identification based on keystroke dynamics, using
classic machine learning models. Our results indicate that practical, real-world 
identification based on keystroke dynamics may be feasible.

\bibliographystyle{plain}

\bibliography{references.bib}

\end{document}

%% file: preamble.tex
\usepackage{amsmath,amsthm, amsfonts, amssymb, amsxtra, amsopn}
\usepackage{pgfplots}
\pgfplotsset{compat=1.13}
\usepackage{pgfplotstable}
\usepackage{graphicx,grffile}
\usepackage{multirow}
\usepackage{booktabs}
\usepackage{listings}
\usepackage{cmap}
\usepackage{colortbl}
\usepackage{adjustbox}
\usepackage{epsfig}

\usepackage[tableposition=top,font=small,skip=5pt]{caption}
\usepackage{subcaption}
\usepackage{makecell} 

\pgfkeys{
    /pgf/number format/precision=2, 
    /pgf/number format/fixed zerofill=true }
    
\pgfplotstableset{
    /color cells/min/.initial=0,
    /color cells/max/.initial=1000,
    /color cells/textcolor/.initial=,
    color cells/.code={%
        \pgfqkeys{/color cells}{#1}%
        \pgfkeysalso{%
            postproc cell content/.code={%
                \begingroup
                \pgfkeysgetvalue{/pgfplots/table/@preprocessed cell content}\value
\ifx\value\empty
\endgroup
\else
                \pgfmathfloatparsenumber{\value}%
                \pgfmathfloattofixed{\pgfmathresult}%
                \let\value=\pgfmathresult
                \pgfplotscolormapaccess
                    [\pgfkeysvalueof{/color cells/min}:\pgfkeysvalueof{/color cells/max}]%
                    {\value}%
                    {\pgfkeysvalueof{/pgfplots/colormap name}}%
                \pgfkeysgetvalue{/pgfplots/table/@cell content}\typesetvalue
                \pgfkeysgetvalue{/color cells/textcolor}\textcolorvalue
                \toks0=\expandafter{}%
                \xdef\temp{%
                    \noexpand\pgfkeysalso{%
                        @cell content={%
                            \noexpand\cellcolor[rgb]{\pgfmathresult}%
                            \noexpand\definecolor{mapped color}{rgb}{\pgfmathresult}%
                            \ifx\textcolorvalue\empty
                            \else
                                \noexpand\color{\textcolorvalue}%
                            \fi
                            \the\toks0 %
                        }%
                    }%
                }%
                \endgroup
                \temp
\fi
            }%
        }%
    }
}

\PassOptionsToPackage{hyphens}{url}

\usepackage{hyperref}
\hypersetup{colorlinks=true,linkcolor=black,citecolor=black,urlcolor=blue,filecolor=black}
\hypersetup{pdfpagemode=UseNone,pdfstartview=}

\usepackage{enumitem}
\setlist[itemize]{noitemsep, topsep=0pt}

\advance\oddsidemargin by -0.35in
\advance\textwidth by 0.7in

\advance\topmargin by -0.4in
\advance\textheight by 0.8in

\long\def\symbolfootnotetext[#1]#2{\begingroup%
\def\thefootnote{\fnsymbol{footnote}}\footnotetext[#1]{#2}\endgroup}

\DeclareMathOperator{\thth}{th}

\let\vvv=\v
\def\v{{\tt v}}

\def\x{{\tt x}}
\def\y{{\tt y}}

%% file: figures/kdFeatures.tex
    \begin{tikzpicture}[scale=0.8]
    
    \draw[red,thick] plot[] coordinates {
    (-0.5,0.5) (0.0,0.5) (0.0,1.5) (1.5,1.5) (1.5,0.5) (3.0,0.5) (3.0,1.5) (4.5,1.5) (4.5,0.5)
    (5.5,0.5) (5.5,1.5) (7.0,1.5) (7.0,0.5) (8.5,0.5) (8.5,1.5) (10.0,1.5) (10.0,0.5) (10.5,0.5)};

    \draw[red,dashed] (1.5,1.5) -- (1.5,6.0);
    \draw[red,dashed] (3.0,1.5) -- (3.0,3.0);
    \draw[red,dashed] (3.0,4.0) -- (3.0,6.0);
    
    \draw[red,dashed] (7.0,1.5) -- (7.0,5.0);
    \draw[red,dashed] (8.5,1.5) -- (8.5,6.0);
    
    \node at (0.75,1.0) {\footnotesize Key~1};
    \node at (2.25,1.3) {\footnotesize Key~1};
    \node at (2.25,0.8) {\footnotesize press};
    \node at (3.75,1.15) {\footnotesize Key~1};
    \node at (3.75,0.725) {\footnotesize release};

    \node at (6.25,1.0) {\footnotesize Key~2};
    \node at (7.75,1.3) {\footnotesize Key~2};
    \node at (7.75,0.8) {\footnotesize press};
    \node at (9.25,1.15) {\footnotesize Key~2};
    \node at (9.25,0.725) {\footnotesize release};

    \draw[ultra thick,color=blue,->] (3.0,5.0) -- (8.5,5.0); 
    \node at (5.75,5.30) {\footnotesize key up-key up};

    \draw[ultra thick,color=blue,->] (3.0,4.0) -- (7.0,4.0); 
    \node at (5.0,4.30) {\footnotesize key up-key down};

    \draw[ultra thick,color=blue,->] (1.5,3.0) -- (7.0,3.0); 
    \node at (4.25,3.30) {\footnotesize key down-key down};

    \draw[ultra thick,color=blue,->] (1.5,2.0) -- (3.0,2.0); 
    \node at (2.25,2.30) {\footnotesize hold};

    \draw[ultra thick,color=blue,->] (7.0,2.0) -- (8.5,2.0); 
    \node at (7.75,2.30) {\footnotesize hold};

    \draw[ultra thick,color=blue,->] (1.5,6.0) -- (8.5,6.0); 
    \node at (5.0,6.3) {\footnotesize total};

    \draw[ultra thick,color=blue,->] (-0.5,0.0) -- (10.5,0.0); 
    \node at (5.0,-0.35) {\footnotesize Time};

     
    \end{tikzpicture}

%% file: figures/keystroke_img.tex
\begin{tikzpicture}[scale=0.25, every node/.style={scale=0.725}]
%
\draw[gray,ultra thick] (7.0,6.0) rectangle (49.0,48.0);
\foreach \x in {7,...,48}{
  \draw[gray,thin] (\x,54-\x) rectangle (\x+1,54-\x+1.0);
}
\draw[green,ultra thick,fill=white] (5.25,4.25) rectangle (47.25,46.25);
\foreach \x in {6,...,47}{
  \foreach \y in {4,...,45}{
    \draw[green,thin] (\x+0.25,\y+1.25) rectangle (\x-0.75,\y+0.25);
  }
}
\draw[brown,ultra thick,fill=white] (3.5,2.5) rectangle (45.5,44.5);
\foreach \x in {6,...,47}{
  \foreach \y in {4,...,45}{
    \draw[brown,thin] (\x-1.5,\y-0.5) rectangle (\x-2.5,\y-1.5);
  }
}
\draw[red,ultra thick,fill=white] (1.75,0.75) rectangle (43.75,42.75);
\foreach \x in {6,...,47}{
  \foreach \y in {4,...,45}{
    \draw[red,thin,opacity=0.5] (\x-3.25,\y-2.25) rectangle (\x-4.25,\y-3.25);
  }
}
\draw[blue,ultra thick,fill=white] (0.0,-1.0) rectangle (42.0,41.0);
\draw[step=1.0,blue,thin] (0.0,-1.0) grid (42.0,41.0);
\draw[black,ultra thick,fill=black] (10.0,10.0) rectangle (14.0,17.0);
\draw[black,ultra thick,fill=black] (18.0,6.0) rectangle (22.0,10.0);
\draw[black,ultra thick,fill=black] (30.0,25.0) rectangle (36.0,29.0);
\node[color=white] at (33.0,27.0) {\Large cutout};
\node at (-0.75,40.5) {\tt a};
\node at (-0.75,39.5) {\tt b};
\node at (-0.75,38.5) {\tt c};
\node at (-0.75,37.5) {\tt d};
\node at (-0.75,36.5) {\tt e};
\node at (-0.75,35.5) {\tt f};
\node at (-0.75,34.5) {\tt g};
\node at (-0.75,33.5) {\tt h};
\node at (-0.75,32.5) {\tt i};
\node at (-0.75,31.5) {\tt j};
\node at (-0.75,30.5) {\tt k};
\node at (-0.75,29.5) {\tt l};
\node at (-0.75,28.5) {\tt m};
\node at (-0.75,27.5) {\tt n};
\node at (-0.75,26.5) {\tt o};
\node at (-0.75,25.5) {\tt p};
\node at (-0.75,24.5) {\tt q};
\node at (-0.75,23.5) {\tt r};
\node at (-0.75,22.5) {\tt s};
\node at (-0.75,21.5) {\tt t};
\node at (-0.75,20.5) {\tt u};
\node at (-0.75,19.5) {\tt v};
\node at (-0.75,18.5) {\tt w};
\node at (-0.75,17.5) {\tt x};
\node at (-0.75,16.5) {\tt y};
\node at (-0.75,15.5) {\tt z};
\node at (-0.75,14.5) {\tt 0};
\node at (-0.75,13.5) {\tt 1};
\node at (-0.75,12.5) {\tt 2};
\node at (-0.75,11.5) {\tt 3};
\node at (-0.75,10.5) {\tt 4};
\node at (-0.75,9.5) {\tt 5};
\node at (-0.75,8.5) {\tt 6};
\node at (-0.75,7.5) {\tt 7};
\node at (-0.75,6.5) {\tt 8};
\node at (-0.75,5.5) {\tt 9};
\node at (-1.2,4.5) {tab};
\node at (-1.0,3.5) {l-s};
\node at (-1.0,2.5) {ba};
\node at (-1.0,1.5) {r-s};
\node at (-1.2,0.5) {cap};
\node at (-1.0,-0.5) {sp};
\node at (7.5,48.5) {\smash{\tt a}};
\node at (8.5,48.5) {\smash{\tt b}};
\node at (9.5,48.5) {\smash{\tt c}};
\node at (10.5,48.5) {\smash{\tt d}};
\node at (11.5,48.5) {\smash{\tt e}};
\node at (12.5,48.5) {\smash{\tt f}};
\node at (13.5,48.5) {\smash{\tt g}};
\node at (14.5,48.5) {\smash{\tt h}};
\node at (15.5,48.5) {\smash{\tt i}};
\node at (16.5,48.5) {\smash{\tt j}};
\node at (17.5,48.5) {\smash{\tt k}};
\node at (18.5,48.5) {\smash{\tt l}};
\node at (19.5,48.5) {\smash{\tt m}};
\node at (20.5,48.5) {\smash{\tt n}};
\node at (21.5,48.5) {\smash{\tt o}};
\node at (22.5,48.5) {\smash{\tt p}};
\node at (23.5,48.5) {\smash{\tt q}};
\node at (24.5,48.5) {\smash{\tt r}};
\node at (25.5,48.5) {\smash{\tt s}};
\node at (26.5,48.5) {\smash{\tt t}};
\node at (27.5,48.5) {\smash{\tt u}};
\node at (28.5,48.5) {\smash{\tt v}};
\node at (29.5,48.5) {\smash{\tt w}};
\node at (30.5,48.5) {\smash{\tt x}};
\node at (31.5,48.5) {\smash{\tt y}};
\node at (32.5,48.5) {\smash{\tt z}};
\node at (33.5,48.5) {\smash{\tt 0}};
\node at (34.5,48.5) {\smash{\tt 1}};
\node at (35.5,48.5) {\smash{\tt 2}};
\node at (36.5,48.5) {\smash{\tt 3}};
\node at (37.5,48.5) {\smash{\tt 4}};
\node at (38.5,48.5) {\smash{\tt 5}};
\node at (39.5,48.5) {\smash{\tt 6}};
\node at (40.5,48.5) {\smash{\tt 7}};
\node at (41.5,48.5) {\smash{\tt 8}};
\node at (42.5,48.5) {\smash{\tt 9}};
\node[rotate=90] at (43.5,49.25) {tab};
\node[rotate=90] at (44.5,49.05) {l-s};
\node[rotate=90] at (45.5,49.05) {ba};
\node[rotate=90] at (46.5,49.05) {r-s};
\node[rotate=90] at (47.5,49.3) {cap};
\node[rotate=90] at (48.5,49.035) {sp};
\end{tikzpicture}

%% file: figures/barClusters.tex
\begin{tikzpicture}[scale=0.75, every node/.style={scale=0.9}]
\pgfkeys{/pgf/number format/.cd,1000 sep={}}
\begin{axis}[
        width  = 0.65*\textwidth,
        height = 8.0cm,
        ymin=0,ymax=100,
        ytick={0,20,40,60,80,100},
        major x tick style = transparent,
        ybar=5*\pgflinewidth,
        bar width=38.0pt,
        ylabel = {Count},
        symbolic x coords={Easy, Moderate, Difficult},
        xticklabels={Easy, Moderate, Difficult},
	y tick label style={
    		/pgf/number format/.cd,
   		fixed,
   		fixed zerofill,
    		precision=0},
        xtick = data,
        x tick label style={
		font=\small,
		},
        nodes near coords,
        every node near coord/.append style={
								   /pgf/number format/.cd,
								   fixed,
								   fixed zerofill,
								   precision=0},
        enlarge x limits=0.25,
        legend cell align=left,
        legend style={
                at={(0.915,0.52)},
                anchor=south,
                column sep=1ex
        },
]
\addplot [fill=blue,opacity=1.00]%
coordinates {
(Easy, 15)
(Moderate, 85)
(Difficult, 48)
};
\end{axis}
\end{tikzpicture}

%% file: figures/barRange.tex
\begin{tikzpicture}[scale=0.75, every node/.style={scale=0.9}]
\pgfkeys{/pgf/number format/.cd,1000 sep={}}
\begin{axis}[
        nodes near coords,
        nodes near coords greater equal only/.style={
            /pgf/number format/.cd,
             	fixed,
		fixed zerofill,
		precision=0,
            small value/.style={
                /tikz/coordinate,
            },
            every node near coord/.append style={
                check for small values/.code={
                    \begingroup
                    \pgfkeys{/pgf/fpu}
                    \pgfmathparse{\pgfplotspointmeta<#1}
                    \global\let\result=\pgfmathresult
                    \endgroup
                    %
                    %
                    \pgfmathfloatcreate{1}{1.0}{0}
                    \let\ONE=\pgfmathresult
                    \ifx\result\ONE
                        \pgfkeysalso{/pgfplots/small value}
                    \fi
                },
                check for small values,
            },
        },
        nodes near coords greater equal only=1,
        width  = 0.85*\textwidth,
        height = 8.0cm,
        ymin=0,ymax=114,
        ytick={0,20,40,60,80,100},
        major x tick style = transparent,
        ybar=5*\pgflinewidth,
        bar width=30.0pt,
        xlabel = {Accuracy range},
        ylabel = {Number of users},
        symbolic x coords={0.00,0.05,0.10,0.15,0.20,0.25,0.30,0.35,0.40,0.45,0.50,0.55,0.60,0.65,0.70,0.75,
        		0.80,0.85,0.90,0.95,1.00},
        xticklabels={0.00,,0.10,,0.20,,0.30,,0.40,,0.50,,0.60,,0.70,,0.80,,0.90,,1.00},
	y tick label style={
    		/pgf/number format/.cd,
   		fixed,
   		fixed zerofill,
    		precision=0},
        xtick = data,
        x tick label style={
		font=\small,
		},
        enlarge x limits=0.05,
        legend cell align=left,
        legend style={
                at={(0.915,0.52)},
                anchor=south,
                column sep=1ex
        },
]
\addplot [fill=blue,opacity=1.00]%
coordinates {
(0.05,0)
(0.10,0)
(0.15,0)
(0.20,0)
(0.25,0)
(0.30,0)
(0.35,0)
(0.40,0)
(0.45,0)
(0.50,0)
(0.55,0)
(0.60,0)
(0.65,2)
(0.70,0)
(0.75,10)
(0.80,0)
(0.85,32)
(0.90,0)
(0.95,104)
(1.00,0)
};
\end{axis}
\end{tikzpicture}